\documentclass[10pt,twocolumn,letterpaper]{article}

\usepackage{cvpr}              % To produce the CAMERA-READY version
\usepackage{graphicx}
\usepackage{amsmath}
\usepackage{amssymb}
\usepackage{booktabs}
\usepackage{lscape}
\usepackage[table,xcdraw]{xcolor}
\usepackage{multirow}

\usepackage[pagebackref,breaklinks,colorlinks]{hyperref}

\usepackage[capitalize]{cleveref}
\crefname{section}{Sec.}{Secs.}
\Crefname{section}{Section}{Sections}
\Crefname{table}{Table}{Tables}
\crefname{table}{Tab.}{Tabs.}

\def\cvprPaperID{*****} % *** Enter the CVPR Paper ID here
\def\confName{CVPR}
\def\confYear{2022}

\begin{document}

\title{Exploiting Semantic Role Contextualized Video Features \\ for Multi-Instance Text-Video Retrieval \\ EPIC-KITCHENS-100 Multi-Instance Retrieval Challenge 2022}

\author{Burak Satar$^{1, 2}$\\
\and
Zhu Hongyuan$^{1}$\\
\and
Hanwang Zhang$^{2}$\\
\and
Joo Hwee Lim$^{1, 2}$
\and
$^{1}$Institute for Infocomm Research, A*STAR, Singapore
\and
$^{2}$School of Computer Science and Engineering, NTU, Singapore\\
{\tt\small \{burak\_satar, zhuh, joohwee\}@i2r.a-star.edu.sg}, \tt\small hanwangzhang@ntu.edu.sg}

\maketitle

%%%%%%%%% ABSTRACT
\begin{abstract}
In this report, we present our approach for EPIC-KITCHENS-100 Multi-Instance Retrieval Challenge 2022. We first parse sentences into semantic roles corresponding to verbs and nouns; then utilize self-attentions to exploit semantic role contextualized video features along with textual features via triplet losses in multiple embedding spaces. Our method overpasses the strong baseline in normalized Discounted Cumulative Gain (nDCG), which is more valuable for semantic similarity. Our submission is ranked 3rd for nDCG and ranked 4th for mAP.  
\end{abstract}

\section{Introduction}
\label{sec:intro}
%We present the strategy we used for the EPIC KITCHENS-100 2022 Multi-Instance Retrieval Challenge in this report. 
With the rise of videos uploaded by users via social media channels, cross-modal retrieval of video data and natural language descriptions has gained popularity.
The goal of video-to-text retrieval, given a query action segment, is to rank captions in a gallery set so that those with a higher rank are more semantically related to the video action. Text-to-video retrieval, on the other hand, ranks videos based on a query caption. %\vspace{5pt}

While most methods \cite{Liu2019a, mithun2020, dong_cvpr19}  use one joint embedding space to align video and text features, recent methods \cite{Chen_2020_CVPR, wray2019fine} use multiple embedding spaces to match video features into the noun and verb embedding spaces along with the textual features but did not consider their interactions. Moreover, it is relatively easy to parse the textual features into verb and noun levels since an off-the-shelf toolkit could be used. However, mapping a video feature into the object and action levels is still challenging, which corresponds to noun and verb levels in text. %\vspace{5pt}

Inspired by \cite{satar_2021, satar2022rome, gabeur2020mmt}, we implement self-attentions to exploit visual features on top of the baseline, JPoSE \cite{wray2019fine} by leveraging the contexts from nouns and verbs of the text query with details in the following section. While we outperform the strong baseline in normalized Discounted Cumulative Gain (nDCG), which is more beneficial for semantic similarity, we fall short in mean Average Precision (mAP), a traditional technique for binary relevance. Our approach is ranked third for nDCG, while it is ranked fourth for mAP. We also analyze various failure examples to save the time of the following researchers on this task.

\section{Method}
\label{sec:method}
We follow the baseline work \cite{wray2019fine}: we create a pair of functions that map videos and texts into a joint embedding space, in which embeddings for matched texts and videos should be close together, and embeddings for mismatched texts and videos should be far apart, given a video and a query text. A suitable embedding space should also ensure that related videos/texts stay close together. 

\begin{figure*}[htb]
\centering
\centerline{\includegraphics[width=13.5cm]{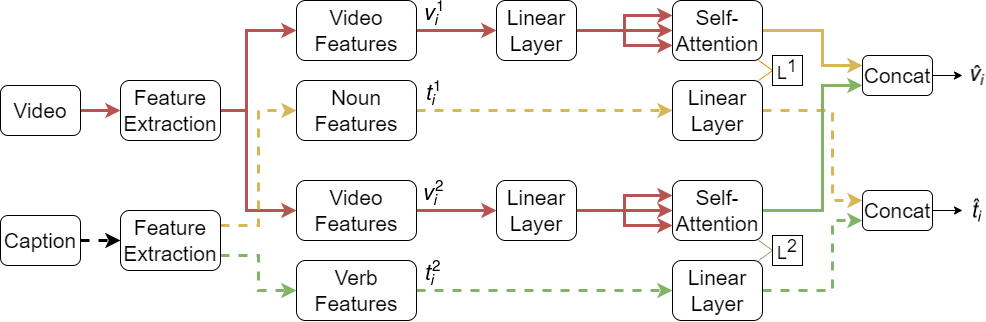}}
\caption{We first parse caption into the noun $t_{i}^{1}$ and verb $t_{i}^{2}$ levels, followed by linear layers. We utilize linear layers to embed corresponding video features $v_{i}^{1}$, $v_{i}^{2}$ and use a self-attention layer to exploit contextualized features.
Then, we concatenate textual and visual features to compute the distance between these representations, $\hat{v_{i}}$ and $\hat{t_{i}}$. L1 and L2 refer to triplet losses. }\vspace{-10pt}
\label{fig:model}
\end{figure*}

With this motivation, we first parse caption into the noun $t_{i}^{1}$ and verb $t_{i}^{2}$ levels, followed by linear layers. We utilize linear layers to embed corresponding video features $v_{i}^{1}$, $v_{i}^{2}$ and use a self-attention layer to exploit contextualized features.
Then, we concatenate textual and visual features to compute the distance between these representations, $\hat{v_{i}}$ and $\hat{t_{i}}$. L1 and L2 refer to triplet losses. The more details of the loss functions are in Eq. \ref{loss} and baseline paper \cite{wray2019fine} and the architecture details are in Fig. \ref{fig:model}

In Eq. \ref{loss},  the first two rows refer to cross-modal losses, and the last two rows indicate within-modal losses. $\theta$ function denotes two fully connected layers. $\delta$ function signifies two linear layers and one self-attention layer. $m$ refers to the constant margin, while $d$ is the distance function. While $i$ refers to the selected video, $j$ and $k$ denote positive and negative samples, respectively.

\begin{equation}
\begin{split}
    L & = \lambda_{v,t} \sum_{i, j, k} \max(0, d(\delta_{v_{i}},\theta_{t_{j}}) - d(\delta_{v_{i}}, \theta_{t_{k}}) + m) \\
    & + \lambda_{t,v} \sum_{i, j, k} \max(0, d(\theta_{t_{i}}, \delta_{v_{j}}) - d(\theta_{t_{i}}, \delta_{v_{k}}) + m) \\
     & + \lambda_{v,v} \sum_{i, j, k} \max(0, d(\delta_{v_{i}},\delta_{v_{j}}) - d(\delta_{v_{i}}, \delta_{v_{k}}) + m) \\
    & + \lambda_{t,t} \sum_{i, j, k} \max(0, d(\theta_{t_{i}},\theta_{t_{j}}) - d(\theta_{t_{i}}, \theta_{t_{k}}) + m)
\end{split}
\label{loss}
\end{equation}

Eq. \ref{encoder} shows that the visual features $V_i$ are fed into the self-attention layer to encode into $z_s$. Then, a feed-forward layer FF outputs the final contextualized appearance feature. Normalization of the layer is done under the Norm function. 

\begin{equation}
\begin{aligned}
&z_s = \textrm{Norm}\big(\textrm{MultiHead}(V_i, V_i, V_i) + V_i\big)& \\
&E_v = \textrm{Norm}\big(\textrm{FF}(z_s) + z_s\big)& \\
\label{encoder}
\end{aligned}  
\end{equation}

For multi-headed attention layers, we follow \cite{NIPS2017_3f5ee243}, as formulated in Eq. \ref{transformer}. All of the W matrices are learned during the training procedure. Since it is a self-attention layer, query Q is the same as key K and value V. After each attention layer, layer normalization and the residual connection are implemented.

\begin{equation}
\begin{aligned}
&\textrm{MultiHead}(Q,K,V) = \textrm{Concat}(\textrm{Head}_1, ..., \textrm{Head}_h)W^O \\ 
&\textrm{Head}_i = \textrm{Attention}(QW_i^Q, KW_i^K, VW_i^V) \\ 
&\textrm{Attention}(Q,K,V) = \sigma\Big(\frac{QK^T}{\sqrt{d}}V\Big)
\end{aligned}
\label{transformer}
\end{equation}

\begin{table*}[!htb]
\centering
\caption{Multi-instance retrieval results on the EPIC-KITCHENS-100 test split. T2V and V2T stand for Text-to-Video and Video-to-Text retrieval, respectively. While the above part of the table compares with baselines, the lower part shares the result of this year's competition.}
\label{result_test}
\resizebox{\textwidth}{!}{%
\begin{tabular}{|ccccccc|}
\hline
\multicolumn{7}{|c|}{\cellcolor[HTML]{FFCE93}\textbf{Comparison to Baselines}} \\ \hline
\multicolumn{1}{|c|}{} &
  \multicolumn{3}{c|}{\textbf{mean Average Precision  (mAP)}} &
  \multicolumn{3}{c|}{\textbf{normalised Discounted  Cumulative Gain (nDCG)}} \\ \cline{2-7} 
\multicolumn{1}{|c|}{\multirow{-2}{*}{\textbf{Method}}} &
  Average &
  T2V &
  \multicolumn{1}{c|}{V2T} &
  Average &
  T2V &
  V2T \\ \hline
\multicolumn{1}{|c|}{MI-MM} &
  27.58 &
  23.08 &
  \multicolumn{1}{c|}{32.09} &
  42.10 &
  40.48 &
  43.72 \\ \hline
\multicolumn{1}{|c|}{JPoSE*} &
  43.95 &
  38.18 &
  \multicolumn{1}{c|}{49.71} &
  53.40 &
  51.60 &
  55.21 \\ \hline
\multicolumn{1}{|c|}{JPoSE \cite{wray2019fine}} &
  44.01 &
  38.11 &
  \multicolumn{1}{c|}{49.91} &
  53.53 &
  51.55 &
  55.51 \\ \hline
\multicolumn{1}{|c|}{DCRL \cite{Xiaoshuai}} &
  \textbf{44.23} &
  \textbf{38.49} &
  \multicolumn{1}{c|}{\textbf{49.96}} &
  53.56 &
  51.83 &
  55.28 \\ \hline
\multicolumn{1}{|c|}{Our Method} &
  42.81 &
  38.10 &
  \multicolumn{1}{c|}{47.52} &
  \textbf{55.33} &
  \textbf{54.12} &
  \textbf{56.55} \\ \hline
\multicolumn{7}{|c|}{\cellcolor[HTML]{FFCE93}\textbf{Comparison to Other Users}} \\ \hline
\multicolumn{1}{|c|}{} &
  \multicolumn{3}{c|}{\textbf{mean Average Precision  (mAP)}} &
  \multicolumn{3}{c|}{\textbf{normalised Discounted  Cumulative Gain (nDCG)}} \\ \cline{2-7} 
\multicolumn{1}{|c|}{\multirow{-2}{*}{\textbf{User}}} &
  Average &
  T2V &
  \multicolumn{1}{c|}{V2T} &
  Average &
  T2V &
  V2T \\ \hline
\multicolumn{1}{|c|}{haoxiaoshuai} &
  44.02 (3) &
  38.34 (3) &
  \multicolumn{1}{c|}{49.69 (3)} &
  53.06 (4) &
  51.31 (4) &
  54.82 (4) \\ \hline
\multicolumn{1}{|c|}{Our Method} &
  42.81 (4) &
  38.10 (4) &
  \multicolumn{1}{c|}{47.52 (4)} &
  55.33 (3) &
  54.12 (3) &
  56.55 (3) \\ \hline
\multicolumn{1}{|c|}{afalcon} &
  49.77 (1) &
  44.39 (1) &
  \multicolumn{1}{c|}{55.15 (1)} &
  61.02 (2) &
  58.88 (2) &
  63.16 (2) \\ \hline
\multicolumn{1}{|c|}{kevin.lin} &
  47.39 (2) &
  40.95 (2) &
  \multicolumn{1}{c|}{53.84 (2)} &
  61.44 (1) &
  59.60 (1) &
  63.29 (1) \\ \hline
\end{tabular}%
}
\end{table*}

\section{Experiments}

\subsection{Implementation Details}
While $\lambda_{t,v}$ equals 2.0, the other constant margins equal 1.0. The batch size is 64, and the learning rate is 0.01.

\textbf{Dataset.} We undertake experiments on the EPIC-KITCHENS-100 dataset \cite{Damen2021RESCALING}, which is a collection of unscripted egocentric action data across the world, to demonstrate the efficiency of our strategy.

\textbf{Features.} We use the video features extracted by TBN \cite{Kazakos_2019_ICCV}. Each one is an nx25x1024 matrix holding a python dictionary containing the RGB, flow, and audio features, where n is the number of video clips. The number of training and test set pairs is 67217 and 9668, respectively. Each feature is followed by temporal mean pooling, making the shape nx1x1024. We utilize the textual features given by \cite{wray2019fine} using a Word2Vec model trained on the Wikipedia corpus. spaCy parser \cite{eng_spacy2} is used to disentangle the text caption into different PoS tags. The model is trained with the default values of the baseline. 

\textbf{Evaluation Metrics.} We utilize two assessment metrics, mAP and nDCG, on the test set to evaluate submissions for action retrieval. Mean Average Precision (mAP) was employed for retrieval baselines because it allows the whole ranking to be analyzed on binary relevance. nDCG has already been used to retrieve information \cite{wray2021semantic}. It necessitates the use of similarity scores throughout the entire test set. 

\subsection{Results}

Table \ref{result_test} shows the comparison between our method and the baselines. It also compares with the methods attended to this year's challenge. While our method overpass all the baselines on nDCG, it falls short on mAP. The MI-MM approach projects both modalities onto a shared action space using linear layers via max-margin loss, which is a simplified version of \cite{miech20endtoend}. The JPoSE approach \cite{wray2019fine} uses a triplet loss to separate captions into the verb and noun spaces. The JPoSE* refers to our implementation. DCRL \cite{Xiaoshuai} considers both inter-modal and intra-modal constraints at the same time to retain both cross-modal semantic similarity and modality-specific consistency in the embedding space.

\textbf{Failure cases.}
We also share failure cases which could be helpful for other researchers. For every experiment, we give the results approximately compared to baseline JPoSE \cite{wray2019fine}. 1) If we implement self-attention to the textual features as it is done to the video features, the results decrease around 2-3\%. 2) When we increase the batch size or embedding size, the results decrease 1-2\%. 3) We get 1-2\% lower results when applying temporal max-pooling rather than mean pooling.

\section{Conclusion}

In this report, we propose an approach to exploit contextualized video features via self-attentions and disentangling them into multiple embedding spaces. It also parses text into corresponding embedding spaces, and then the similarity between representations is calculated via triplet loss. While our strategy outperforms the strong baseline in normalized Discounted Cumulative Gain (nDCG), a semantic similarity measurement, it falls short in mean Average Precision (mAP), a standard measure of binary relevance. For nDCG, our proposal is ranked third, and for mAP, it is ranked fourth. We plan to exploit each video feature separately via novel fusion methods as well as utilize domain-specific features such as hand-object relations for future work.

{\small
\bibliographystyle{ieee_fullname}
\bibliography{egbib}
}
\end{document}